%% file: main.tex
\documentclass[sigconf]{acmart}

\usepackage{pifont}
\usepackage[most]{tcolorbox}
\usepackage{graphicx}
\usepackage{subcaption}
\usepackage{multirow} 
\usepackage{enumitem}
\usepackage{soul}
\usepackage{xspace}
\AtBeginDocument{%
  }

\copyrightyear{2026}
\acmYear{2026}
\setcopyright{cc}
\setcctype{by}
\acmConference[KDD '26]{Proceedings of the 32nd ACM SIGKDD Conference on Knowledge Discovery and Data Mining V.2}{August 09--13, 2026}{Jeju Island, Republic of Korea}
\acmBooktitle{Proceedings of the 32nd ACM SIGKDD Conference on Knowledge Discovery and Data Mining V.2 (KDD '26), August 09--13, 2026, Jeju Island, Republic of Korea}
\acmDOI{10.1145/3770855.3817570}
\acmISBN{979-8-4007-2259-2/2026/08}

\newcommand{\DS}{{\textsc{MKG-RAG-Bench}\xspace}}
\newcommand{\name}{{\textsc{MKG-RAG}\xspace}}

\begin{document}

\title{{\DS}: Benchmarking Retrieval in Multimodal Knowledge Graph–Augmented Generation}


\author{Xiaochen Wang}
\affiliation{%
  \institution{The Pennsylvania State University}
  \city{University Park}
  \state{Pennsylvania}
  \country{USA}}
\email{xcwang@psu.edu}

\author{Bao Hoang}
\affiliation{%
  \institution{Michigan State University}
  \city{East Lansing}
  \state{Michigan}
  \country{USA}
  }
\email{hoangbao@msu.edu}

\author{Han Liu}
\affiliation{%
  \institution{Dalian University of Technology}
  \city{Dalian}
  \state{Liaoning}
  \country{China}
}
\email{liu.han.dut@gmail.com}

\author{Ting Wang}
\affiliation{%
  \institution{Stony Brook University}
  \city{Stony Brook}
  \state{New York}
  \country{USA}
}
\email{wang@cs.stonybrook.edu}

\author{Fenglong Ma}
\affiliation{%
  \institution{The Pennsylvania State University}
  \city{University Park}
  \state{Pennsylvania}
  \country{USA}
}
\email{fenglong@psu.edu}
\authornote{Corresponding author.}

\renewcommand{\shortauthors}{Xiaochen Wang, Bao Hoang, Han Liu, Ting Wang, and Fenglong Ma}
\begin{abstract}
Retrieval-augmented generation (RAG) over knowledge graphs has emerged as a promising approach for grounding large language models, yet existing benchmarks largely overlook the challenges of retrieval in multimodal knowledge graph RAG (MKG-RAG). In practice, retrieval is a critical bottleneck: multimodal knowledge is heterogeneous, difficult to align across modalities, and often poorly served by retrievers designed for unstructured corpora. To address this gap, we introduce ~{\DS}, a cross-domain benchmark explicitly designed to evaluate retrieval in MKG-RAG. ~{\DS} is constructed from two multimodal knowledge graphs spanning general and medical domains, and includes carefully aligned question-answering datasets that support controlled evaluation of both retrieval and downstream generation. The benchmark is built using an LLM-based curation pipeline that filters low-utility knowledge, generates structurally grounded queries with exact supervision, and systematically covers diverse modality configurations. Through extensive experiments across representative retriever families and modality settings, we show that effective multimodal retrieval remains challenging yet crucial for end-to-end MKG-RAG performance, and that retrieval quality strongly determines generation outcomes. By isolating retrieval as a first-class evaluation target, {\DS} provides a principled foundation for diagnosing current limitations and advancing multimodal knowledge graph RAG systems.\footnote{The benchmark can be accessed via: \url{https://github.com/XiaochenWang-PSU/MKG-RAG-Bench}.}
\end{abstract}



\maketitle

\input{tex/intro-final}

\input{tex/preliminaries-final}
\input{tex/curation-final}
\input{tex/experiments-final}

\input{tex/conclusion}

\section*{Acknowledgements}
This research was partially supported by a 2025/2026 Rising Researcher Grant from Penn State's Institute for Computational \& Data Sciences (RRID:SCR\_025154) and the National Science Foundation under Grant No. 2333790 and 2238275.

\bibliographystyle{ACM-Reference-Format}
\bibliography{sample-base}

\appendix
\input{tex/appendix}

\end{document}

%% file: tex/intro-final.tex
\section{Introduction}
\label{sec:intro}

Retrieval-augmented generation (RAG) enhances large language models (LLMs) by retrieving information relevant to a given query from external sources and conditioning generation on this retrieved evidence~\cite{lewis2020retrieval}. While RAG has proven effective in improving factual grounding, conventional approaches predominantly retrieve from unstructured textual corpora that are noisy, fragmented, and weakly connected. These limitations often hinder reliable evidence selection and multi-step reasoning, especially for complex queries.
To address these issues, \textit{knowledge graph–based RAG} (KG-RAG) has been proposed~\cite{edge2024local}. By retrieving information from structured knowledge graphs, KG-RAG enables more coherent, contextualized, and comprehensive knowledge access, leading to improved answer quality and faithfulness~\cite{wang2026gpr}. However, existing KG-RAG methods are largely confined to \textit{textual knowledge graphs}, which substantially limits their applicability in real-world scenarios where critical information is conveyed through images, charts, tables, and other non-textual modalities.

\begin{figure*}[t]
  \centering
  \includegraphics[width=0.95\linewidth]{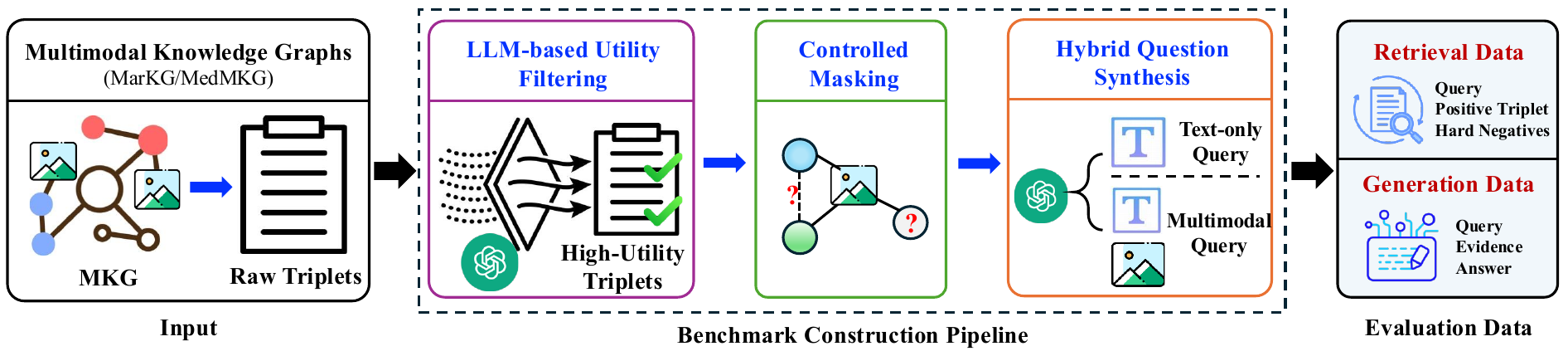}
  \vspace{-0.15in}
  \caption{The proposed pipeline for benchmark construction using multimodal knowledge graphs and LLMs.}
  \label{fig:pipeline}
\end{figure*}

\smallskip
\noindent \textbf{\ul{Multimodal KG-RAG.}}
To overcome this limitation, we explore the task, {multimodal KG-RAG} (\name), which integrates structured knowledge across multiple modalities to support more expressive retrieval and reasoning. Formally, given a query $q$, a multimodal knowledge graph $\mathcal{G}$, a retriever $\textsf{Retriever}$, and a multimodal large language model $\textsf{MLLM}$, the goal of {\name} is to generate a response $\mathcal{R}$, i.e.,
\[
\mathcal{R} = \textsf{MLLM}(q, \textsf{Retriever}(q, \mathcal{G})).
\]
The most closely related setting is multimodal RAG
~\cite{chen2022murag}, which follows the traditional RAG pipeline by retrieving multimodal content from unstructured external corpora (e.g., Wikipedia) rather than from a structured multimodal knowledge graph. To the best of our knowledge, \textbf{no existing benchmark or prior work} systematically studies retrieval and generation in the {\name} setting.

\smallskip
\noindent \textbf{\ul{Motivation and Challenges.}}
Multimodal KG-RAG is benchmark-worthy because it evaluates a model’s ability to retrieve, align, and reason over \emph{structured multimodal knowledge}, a critical capability for real-world systems that is not captured by existing RAG, KG-RAG, or multimodal benchmarks. This work aims to introduce the \textbf{first} benchmark for the {\name} task, consisting of curated datasets and evaluation protocols that explicitly target retrieval quality and multimodal grounding. Designing such a benchmark, however, poses several non-trivial challenges:
\begin{itemize}[leftmargin=*]
    \item \textbf{Heterogeneous Retrieval.}
    Unlike conventional knowledge graphs, where both queries and target triplets are unimodal, multimodal knowledge graphs support heterogeneous modality combinations across queries and retrieval targets. In {\name}, queries may be text-only, image-only, or multimodal, while relevant graph components may combine textual, visual, and numerical information in different ways. These combinations induce distinct retrieval behaviors and demand different modeling strategies. A meaningful {\name} benchmark must therefore systematically cover diverse multimodal retrieval scenarios to reflect realistic usage and to enable fair comparison among retrievers.

    \item \textbf{Multimodal Knowledge Graph--Query Alignment.}
    In {\name}, relevant knowledge may not be explicitly aligned with the query at the surface level, particularly when the query and target nodes differ in modality or abstraction. For example, a textual query may require grounding in visual evidence, while an image-based query may necessitate retrieving semantically related textual entities and relations. Addressing such cross-modal and cross-structural alignment requires joint reasoning over graph topology, semantic relationships, and modality-specific representations. Accordingly, a {\name} benchmark must include queries that demand non-trivial alignment beyond shallow modality matching.


    \item \textbf{Benchmark Utility.}
    The primary goal of the benchmark is to evaluate both the effectiveness of multimodal retrieval and the grounding quality of multimodal LLMs. This requires careful construction to ensure that the retrieved knowledge is informative and well-aligned with downstream queries. Multimodal knowledge graphs often contain generic or weakly informative triplets, which can lead to ambiguous or poorly grounded questions and obscure the contribution of retrieval. A useful {\name} benchmark must prioritize high-utility triplets and disentangle improvements due to effective retrieval from those arising from generative priors.

\end{itemize}

\smallskip
\noindent \textbf{\ul{Our Solution.}}
To address these challenges, we propose a cross-domain, retrieval-oriented {\name} benchmark, named {\DS}, comprising two multimodal knowledge graphs and corresponding multimodal question-answering datasets. The benchmark is designed to enable comprehensive evaluation across both the retrieval and generation stages of multimodal KG-RAG. The benchmark construction pipeline is shown in Figure~\ref{fig:pipeline}. Specifically, we use an LLM to select high-utility triplets and employ heuristic strategies to ensure strong alignment between the selected triplets and the constructed queries. The same LLM, with different prompts, is further leveraged to generate multimodal, knowledge-grounded questions across diverse retrieval scenarios, thereby systematically covering heterogeneous modality combinations. We conduct extensive experiments with representative retriever architectures across different multimodal settings, evaluating both effectiveness and efficiency throughout the retrieval and generation pipeline. \textit{While this work primarily focuses on evaluating {\name} using \textbf{training-free} approaches, the benchmark is also applicable to training-based methods, including retriever training and MLLM fine-tuning.}

\begin{table*}[t]
\centering
\caption{Comparison of representative multimodal RAG and KG-RAG benchmarks. }
\label{tab:benchmarks_features}
\vspace{-0.15in}
\resizebox{1\textwidth}{!}{%
\begin{tabular}{l|cccccc}
\toprule
\textbf{\shortstack{Benchmark\\Name}} &
\textbf{\shortstack{Knowledge Source \\Type}} &
\textbf{\shortstack{Knowledge \\Structure}} &
\textbf{\shortstack{Retrieval\\Unit}} &
\textbf{\shortstack{Heterogeneous\\Query Support}} &
\textbf{\shortstack{Evaluation\\Granularity}} &
\textbf{\shortstack{Cross-Domain\\Coverage}} \\
\midrule
OK-VQA~\cite{marino2019ok}              &  Web & Unstructured & Passage & Multimodal & Generation-only & Single-domain\\
M$^2$RAG~\cite{ma2024multi}         & Web & Unstructured & Passage and Image & Multimodal & Generation-only & Single-domain \\
CRAG-MM~\cite{wang2025crag}             & Web & Ad-hoc MKG & Passage and Image & Multimodal & Generation-only & Multi-domain\\

Dyn-VQA~\cite{libenchmarking}             & Web & Unstructured & Passage and Image & Multimodal & Generation-only  & Multi-domain\\\hline

WebQSP~\cite{yih2016value}              & Textual KG & Textual KG & Textual Triplet & Text-only & Generation-only  &Single-domain \\
CWQ~\cite{talmor2018web} & Textual KG & Textual KG & Textual Triplet & Text-only  & Generation-only &Single-domain \\
GrailQA~\cite{gu2021beyond}             & Textual KG & Textual KG & Textual Triplet & Text-only & Generation-only  & Multi-domain\\\hline

REAL-MM-RAG~\cite{wasserman2025real}         & Multimodal Documents & PDF & Chunk & Text-only & Retrieval-only  & Multi-domain\\

DocBench~\cite{zou2025docbench} & Multimodal Documents & PDF & Chunk &Text-only  & Generation-only & Multi-domain \\
MMLongBench~\cite{ma2024mmlongbench} & Multimodal Documents & PDF & Chunk & Text-only  & Generation-only & Multi-domain \\
MMDocRAG~\cite{dong2025benchmarking}            & Multimodal Documents & PDF & Chunk & Text-only & Retrieval \& Generation  & Multi-domain\\
\midrule
\textbf{{\DS} } & Multimodal KG & Explicit MKG & Multimodal Triplet & Multimodal & Retrieval \& Generation & Multi-domain\\
\bottomrule
\end{tabular}
}
\end{table*}

\smallskip
\noindent \textbf{\ul{Contributions.}}
In summary, our main contributions are as follows:
\begin{itemize}[leftmargin=*]
    \item We identify a critical research gap in {\name}, highlighting that retrieval, despite its central role, is underexplored and insufficiently evaluated in existing multimodal RAG research.
    \item We introduce the first, cross-domain, retrieval-oriented {\name} benchmark comprising two multimodal knowledge graphs and corresponding question-answering datasets, enabling systematic evaluation of retrieval and downstream generation.
    \item We propose a benchmark curation pipeline based on LLMs that directly addresses key challenges in heterogeneous retrieval, multimodal alignment, and benchmark utility.
    \item We develop a unified evaluation framework and conduct comprehensive experiments across diverse multimodal settings, demonstrating the importance of effective multimodal retrieval and enabling future research on more powerful retrievers for {\name}.
\end{itemize}

%% file: tex/preliminaries-final.tex
\section{Preliminaries}
\label{sec:preliminary}

\subsection{Benchmark Comparisons}

The most closely related line of work is multimodal RAG (M-RAG), which focuses on retrieving multimodal content, such as textual passages and images, from large unstructured corpora. Retrieved multimodal evidence can be directly incorporated into the input of multimodal large language models~\cite{ma2024multi, libenchmarking}, or reorganized into ad-hoc multimodal structures to further enhance generation~\cite{wang2025crag}. However, M-RAG follows the conventional RAG paradigm and does not assume access to a structured multimodal knowledge graph. As a result, retrieval is performed over \textit{unstructured corpora}, and any induced structure is transient and query-specific rather than explicitly represented in the knowledge source. 

Another related line of work is knowledge graph–based RAG (KG-RAG), which aims to retrieve relevant triplets or subgraphs from a given knowledge graph to support grounded generation. Existing KG-RAG approaches~\cite{luo2023reasoning, li2024simple}, however, primarily focus on \textit{textual knowledge graphs} and do not explicitly model or retrieve multimodal information. Consequently, they are not designed to evaluate or support retrieval and reasoning over structured multimodal knowledge.


In contrast, {\name} operates over \textit{pre-defined multimodal knowledge graphs}, where entities, relations, and cross-modal connections are explicitly modeled. This fundamentally changes the retrieval problem: instead of selecting isolated multimodal documents, {\name} requires identifying relevant subgraphs that integrate heterogeneous modalities through structured relations. Consequently, existing M-RAG and KG-RAG benchmarks are insufficient for evaluating {\name}, as they do not capture multimodal graph-based retrieval or disentangle retrieval quality from downstream generation. Table~\ref{tab:benchmarks_features} compares representative benchmarks across multiple perspectives, highlighting the distinctive characteristics of {\name}.

\subsection{Retrievers}
\label{sec:pre_baseline}
Although the {\name} task differs from conventional RAG, multimodal RAG, and KG-RAG, many existing retrieval techniques can be directly adapted to this setting. In this work, we categorize commonly used retrievers into four groups based on their underlying retrieval strategies:
\begin{itemize}[leftmargin=*]
    \item \textbf{Text-only retrievers}~\cite{BM25, DPR, SBERT}. 
    Text-only retrievers treat both queries and candidate triplets as plain text and rank candidates using textual similarity, such as sparse lexical matching or dense semantic embeddings, without incorporating visual information.

    \item \textbf{Fusion-based multimodal retrievers}~\cite{chen2022murag, multimodalSurvey}. 
    Fusion-based retrievers encode textual and visual information into a shared representation space and perform retrieval by measuring cross-modal similarity between multimodal queries and multimodal targets.

    \item \textbf{Captioning-based retrievers}~\cite{Gao2022, SelfAdaptive}. 
    Captioning-based retrievers first convert visual content into textual descriptions using an external image captioning model, and then reduce multimodal retrieval to text-only retrieval over the generated captions.

    \item \textbf{Reranking-based retrievers}~\cite{Yan2024, LLMreranker}. 
    Reranking approaches typically adopt a multi-stage pipeline in which a lightweight first-stage retriever filters a candidate set, followed by a more powerful reranker that performs fine-grained relevance estimation and final selection.
\end{itemize}
To ensure fair and controlled comparison, we evaluate all retrievers in a \textit{training-free} setting by directly using their pretrained weights, without fine-tuning on the proposed benchmark.

\subsection{Preliminary Analysis}
\label{sec:existing}

A seemingly straightforward approach to constructing a multimodal knowledge graph RAG benchmark is to directly combine existing multimodal knowledge graphs, such as MMKG~\cite{liu2019mmkg} and TIVA-KG~\cite{10.1145/3581783.3612266}, with modality-relevant downstream tasks, e.g., A-OKVQA~\cite{schwenk2022okvqa} and K-VQA~\cite{shahMYP19}. However, as discussed in Section~\ref{sec:intro}, such a naive combination fails to adequately capture the requirements of the {\name} task and often leads to severe misalignment between the knowledge source, retrieval objectives, and downstream evaluation. In particular, the knowledge required to answer a given query may not be present in the knowledge graph, while irrelevant retrieved content may dominate the model's input, introducing noise and ultimately degrading performance.

\begin{figure}[t]
  \centering
  \includegraphics[width=\linewidth]{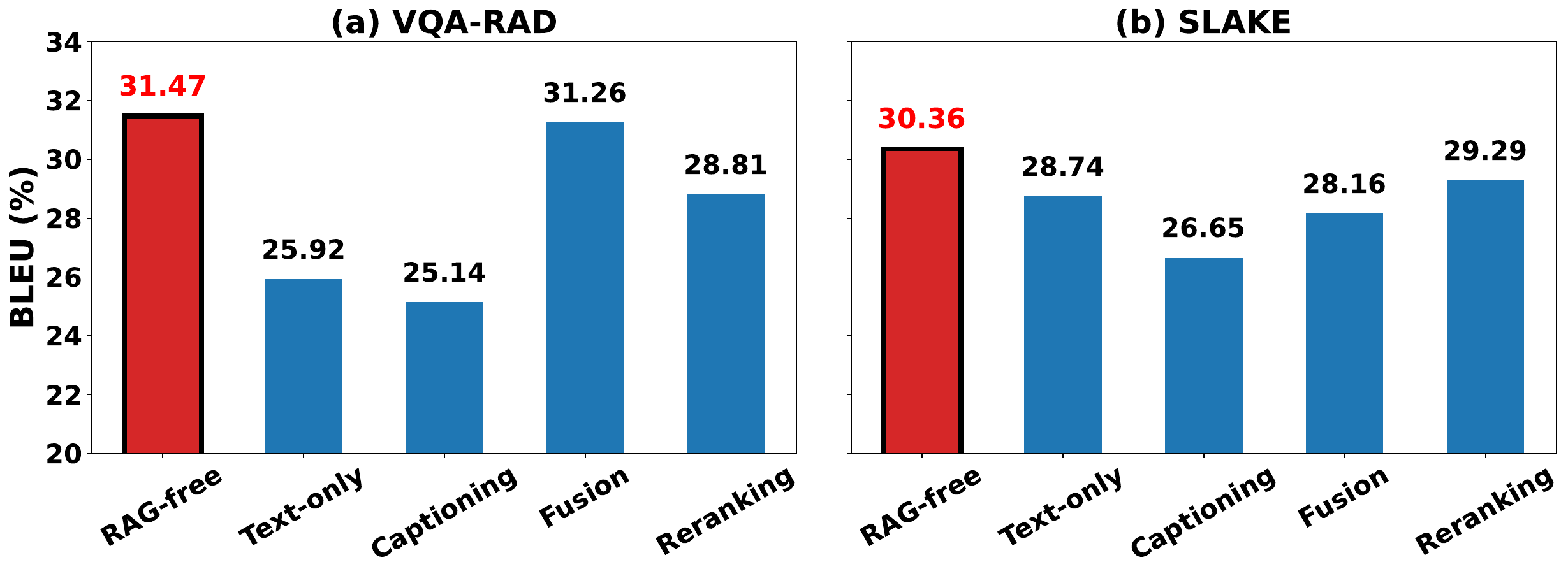}
  \vspace{-0.25in}
  \caption{BLEU (\%) comparison on (a) VQA-RAD and (b) SLAKE datasets. 
  RAG-free consistently achieves the best performance across both benchmarks.}
  \label{fig:pre}
\end{figure}

To empirically illustrate this issue, we conduct a preliminary study using MedMKG~\cite{wang2025medmkgbenchmarkingmedicalknowledge}, a multimodal medical knowledge graph constructed from MIMIC clinical data, to support medical visual question answering on two widely used benchmarks, VQA-RAD~\cite{lau2018dataset} and SLAKE~\cite{liu2021slakesemanticallylabeledknowledgeenhanceddataset}. Specifically, we augment multimodal large language models with retrieval from MedMKG and evaluate their performance under different retrieval strategies. The results are summarized in Figure~\ref{fig:pre}.


Across all experimental settings, models augmented with multimodal retrieval consistently underperform their RAG-free counterparts. Although multimodal retrievers outperform text-only retrievers, they still fail to surpass generation without retrieval augmentation. These results indicate that retrieval from MedMKG does not effectively support the downstream visual question answering tasks. We attribute this behavior to two primary factors: (1) task-critical knowledge required by the MLLM may be absent from the multimodal knowledge graph, and (2) task-irrelevant knowledge present in the graph may be retrieved and injected into the model input, introducing substantial noise that interferes with generation.

Taken together, these observations reveal a fundamental limitation of directly combining existing multimodal knowledge graphs with downstream multimodal tasks for MKG-RAG. The resulting misalignment between knowledge sources, retrieval targets, and evaluation objectives undermines the effectiveness of retrieval augmentation and obscures the true capabilities of multimodal KG-RAG systems. This motivates the need for a dedicated {\name} benchmark that explicitly aligns multimodal knowledge graphs, retrieval objectives, and downstream evaluation—precisely the goal of the benchmark proposed in this work.

%% file: tex/curation-final.tex
\section{Benchmark Construction}
\label{sec:dataset}

\begin{table*}[t]
\centering
\caption{Statistics of {\DS}, where \%MM refers to the percentage of multimodal data.}
\vspace{-0.15in}
\label{tab:mmkg_rt_stats}
\resizebox{1\textwidth}{!}
{%
\begin{tabular}{c|c|ccc|ccc|ccc|cc}
\toprule
\multirow{3}{*}{\textbf{Dataset}} &
\multirow{3}{*}{\textbf{Split}} &
\multicolumn{6}{c|}{\textbf{Retrieval}} &
\multicolumn{5}{c}{\textbf{Generation}} \\
\cmidrule(lr){3-8}\cmidrule(lr){9-13}
&
&\multicolumn{3}{c|}{\textbf{Query}} &
\multicolumn{3}{c|}{\textbf{Triplet}} &
\multicolumn{3}{c|}{\textbf{Question}} &
\multicolumn{2}{c}{\textbf{Answer}} \\
\cmidrule(lr){3-5}\cmidrule(lr){6-8}\cmidrule(lr){9-11}\cmidrule(lr){12-13}
&& \textbf{\# of Queries} & \textbf{\%MM} & \textbf{AvgLen}
& \textbf{\# of Triplets} & \textbf{\%MM} & \textbf{AvgLen}
& \textbf{\# of Questions} & \textbf{\%MM} & \textbf{AvgLen}
& \textbf{\# of Answers} & \textbf{AvgLen} \\
\midrule
\multirow{3}{*}{\textbf{\DS-G}}
&Train & 48,908 & 30.1 & 12.2 & 25,517 & 28.8 & 6.8 & 48,908 & 30.1 & 12.2 & 61,001 & 2.5 \\
& Val & 6,113 & 31.0 & 12.2 & 25,517 & 28.8 & 6.8 & 6,113 & 31.0 & 12.2 & 7,586 & 2.5 \\
&Test & 6,115 & 30.3 & 12.2 & 25,517 & 28.8 & 6.8 & 6,115 & 30.3 & 12.2 & 7,795 & 2.5 \\

\midrule
\multirow{3}{*}{\textbf{\DS-M}}
& Train & 4,781 & 40.6 & 16.2 & 18,468 & 49.0 & 8.7 & 4,781 & 40.6 & 16.2 & 16,564 & 3.7 \\
& Val & 597 & 42.2 & 16.1 & 18,468 & 49.0 & 8.7 & 597 & 42.2 & 16.1 & 2,030 & 3.5 \\
& Test & 599 & 43.1 & 16.2 & 18,468 & 49.0 & 8.7 & 599 & 43.1 & 16.2 & 2,074 & 3.6 \\

\bottomrule
\end{tabular}%
}
\vspace{-0.1in}
\end{table*}

An effective benchmark should not only cover diverse data sources but also support rigorous evaluation across multiple stages of a task. In the proposed {\name} benchmark, we construct datasets from two multimodal knowledge graphs spanning distinct domains: MarKG~\cite{zhang2023multimodalanalogicalreasoningknowledge} for the general domain and MedMKG~\cite{wang2025medmkgbenchmarkingmedicalknowledge} for the medical domain\footnote{The details of these two multimodal knowledge graphs are provided in Appendix~\ref{app:source}.}. This cross-domain design allows us to evaluate the generality of multimodal KG-RAG methods under varying knowledge structures and modality distributions.
To enable faithful evaluation of both retrieval and generation in multimodal KG-RAG, we design a principled data construction pipeline that filters out low-utility knowledge, generates structurally aligned and knowledge-grounded queries with exact supervision, and supports explicit evaluation of both retrieval and generation stages.

\subsection{Dataset Curation}\label{sec:data_curation}
As shown in Figure~\ref{fig:pipeline}, our construction pipeline consists of three stages: (1) LLM-based utility filtering, (2) controlled completeness simulation via triplet masking, and (3) hybrid question synthesis. Each stage is designed to address a specific challenge in benchmarking multimodal KG-RAG.


\smallskip
\noindent \textbf{Step 1: LLM-based Utility Filtering.}
A central challenge in benchmarking multimodal KG-RAG is ensuring that retrieval contributes meaningful and well-grounded evidence to downstream tasks. Multimodal knowledge graphs often contain triplets that are overly generic, definitional, or weakly informative, from which derived queries may be ambiguous or insufficiently grounded, limiting their usefulness for evaluating retrieval.

Given a triplet, we prompt an LLM to assess its \emph{retrieval utility}, i.e., whether it can support a clear and informative information-seeking query. Triplets deemed low-utility are those that tend to yield questions answerable without external evidence or that fail to convey a specific retrieval intent~\cite{yih2016value}. For example, the triplet (\textit{Sun, is\_a, star}) may lead to the question “\textit{What is the Sun?}”, which neither requires retrieval nor benefits from structured knowledge. We remove such triplets from the benchmark to ensure that generated queries are well specified and grounded in retrievable knowledge. This filtering step produces a refined triplet set that better reflects realistic retrieval scenarios in which external knowledge access is necessary for accurate reasoning. The prompt templates and decision criteria are detailed in Appendix~\ref{app:prompts}.

\smallskip
\noindent \textbf{Step 2: Controlled Completeness Simulation.}
To construct queries that require knowledge retrieval while preserving unambiguous supervision, we simulate controlled incompleteness over the retained triplets. Rather than presenting complete knowledge graph facts, we mask one component of each triplet and treat the remaining components as observed context, thereby mimicking information-seeking queries.

For each triplet $(h, r, t)$, we construct two masked variants: (i) \emph{relation masking}, where the relation $r$ is masked and must be inferred from the head and tail, and (ii) \emph{tail masking}, where the tail $t$ is masked and must be inferred from the head and relation. We keep the head entity unchanged as a semantic anchor, which facilitates multimodal grounding when visual information is available and ensures that each query is uniquely supported by its originating knowledge.

We avoid masking the head entity, as doing so would require entity disambiguation beyond the scope of retrieval evaluation and could introduce additional ambiguity. This design allows us to focus on assessing retrievers’ ability to identify relevant relations and entities under realistic multimodal conditions. The resulting masked triplets serve as structurally grounded templates for query generation.

\smallskip
\noindent \textbf{Step 3: Hybrid Question Synthesis.}
Masked triplets provide structured but non-linguistic representations and do not directly resemble natural user queries. To generate realistic questions while preserving exact alignment with the underlying knowledge, we again employ GPT-5~\cite{singh2025openai} to synthesize natural-language questions from masked triplets under two complementary modes.

\textit{Text-only question synthesis.}
For masked triplets derived from text-only knowledge, GPT-5 generates textual questions that explicitly correspond to the missing component. For example, from $(\textit{penicillin}, [\textsc{Mask}], \textit{bacterial infection})$, the LLM generates: ``\textit{What is the relationship between penicillin and bacterial infection?}'' This produces text-only queries whose answers are uniquely grounded in the masked triplet.

\textit{Image-grounded question synthesis.}
For masked triplets whose head entity is associated with an image, we provide the image to GPT-5 and require the head entity to be referenced as \texttt{[Image]}. This enforces explicit visual grounding and prevents trivial lexical matching on entity names. For example, $(\textit{Eiffel Tower}, \textit{located\_in}, [\textsc{Mask}])$ yields the question: `` \includegraphics[height=1.em]{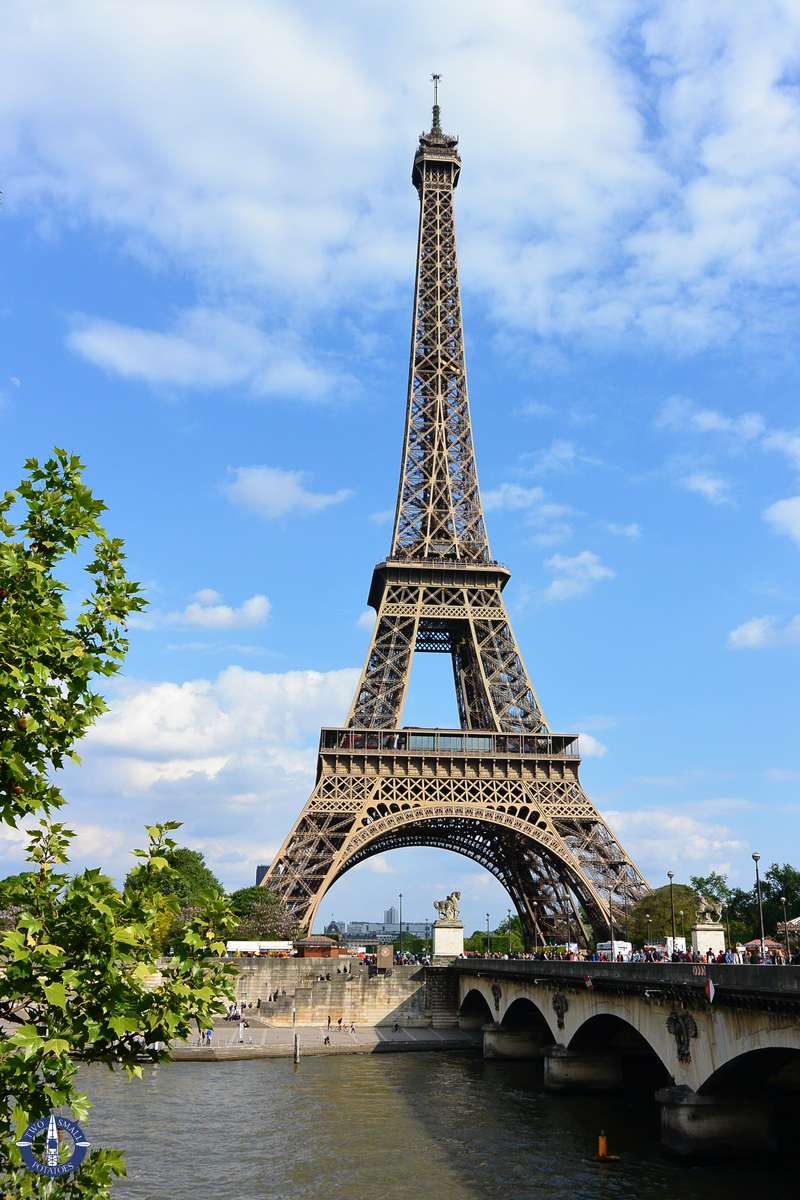}\textit{ Which city is the landmark in the image located in?}'' To avoid shortcut retrieval caused by identical images appearing in both queries and knowledge graph nodes, we apply image augmentations to the question images in Appendix~\ref{app:image}.

Across both modes, question synthesis preserves the semantics of the masked component while producing natural, modality-appropriate queries, enabling retrieval evaluation under both text-only and multimodal settings.

\subsection{Downstream Task Data}

To enable explicit and fine-grained evaluation of retrieval performance, we construct \textbf{retrieval supervision} directly from the aligned query–triplet pairs. For each synthesized question, the originating triplet is treated as the primary positive target. 
Building on the retrieval dataset, we construct \textbf{downstream generation data} in a supervision-preserving manner. Because each question is derived from a masked triplet, the masked component corresponds exactly to the missing information requested by the query. We therefore treat the masked entity or relation as the ground-truth answer.
This process yields question–answer (QA) pairs that are guaranteed to be supported by the aligned multimodal knowledge graph. The resulting QA data includes both text-only and visually grounded instances, reflecting the mixed-modality nature of {\name}. By construction, successful generation requires both correct retrieval and faithful grounding, enabling meaningful end-to-end evaluation. An example of our benchmark construction is provided in Appendix~\ref{app:cons_example}.

\vspace{-0,05in}
\subsection{Dataset Statistics}
\label{sec:ds_stats}
After curating both retrieval and generation datasets from two domain-specific multimodal knowledge graphs, we split the datasets for each domain and each stage into train/validation/test sets with an $8{:}1{:}1$ ratio. The split is performed by questions, so the same question is never shared across train/validation/test sets, while keeping the corpus fixed, following standard IR evaluation.
This split supports not only comprehensive evaluation but also retriever training when needed. To reflect realistic MKG-RAG usage, each split contains a mixture of text-only and multimodal query--triplet pairs, rather than separating modalities into disjoint partitions. As a result, we obtain ~{\DS}, a cross-domain multimodal benchmark that supports training and evaluation under diverse settings in practical MKG-RAG scenarios.
Table~\ref{tab:mmkg_rt_stats} reports the key statistics of {\DS}, where {\DS-G} represents the data obtained from the general multimodal knowledge graph MarKG, and {\DS-M} is the dataset constructed based on the medical multimodal knowledge graph MedMKG.

A notable difference between the two subsets lies in the query--triplet cardinality.
For {\DS-G}, the number of queries notably exceeds the number of triplets.
In contrast, {\DS-M} contains substantially fewer queries than triplets.
This gap is primarily driven by the structural properties of the underlying medical KG, which exhibits more frequent $1{:}n$ patterns.
Popular clinical entities (e.g., \textit{lung} or \textit{pneumonia}) often connect to many other concepts through the same relation, so a fixed incomplete form such as $(h,r,[Mask])$ may correspond to multiple valid tails.
Consequently, many triplets share the same masked form and yield identical queries, resulting in fewer unique queries despite a larger triplet set.

By contrast, the general-domain KG is relatively sparse: relations between everyday entities are less densely connected and tend to produce more unique incomplete forms, leading to a higher diversity of queries per triplet.


%% file: tex/experiments-final.tex
\begin{table*}[t]
\centering
\caption{Retrieval performance (\%) on ~{\DS-G} dataset under different retrieval settings.}
\vspace{-0.1in}
\label{tab:markg_retrieval}
\resizebox{1\textwidth}{!}{%
\begin{tabular}{c|c|l|ccccc|ccccc|ccccc}
\toprule
\multirow{2}{*}{\textbf{\shortstack{Query\\Type}}}
& \multirow{2}{*}{\textbf{\shortstack{Triplet\\Type}}}
& \multirow{2}{*}{\textbf{Method}}
& \multicolumn{5}{c|}{\textbf{NDCG @$K$ $\uparrow$}}
& \multicolumn{5}{c|}{\textbf{Precision @$K$ $\uparrow$}}
& \multicolumn{5}{c}{\textbf{Recall @$K$ $\uparrow$}} \\
\cline{4-18}
& & & 5 & 10 & 20 & 50 & 100
& 5 & 10 & 20 & 50 & 100
& 5 & 10 & 20 & 50 & 100 \\
\midrule

\multirow{5}{*}{\textbf{All}}
& \multirow{5}{*}{\textbf{All}}
& Random & 0.01 & 0.02 & 0.03 & 0.04 & 0.07 & 0.00 & 0.00 & 0.00 & 0.00 & 0.00 & 0.01 & 0.03 & 0.07 & 0.14 & 0.29 \\
& &Text-only & 35.40 & 37.59 & 38.89 & 39.87 & 40.37 & 9.94 & 5.81 & 3.22 & 1.41 & 0.74 & 43.52 & 49.93 & 54.79 & 59.45 & 62.40 \\
& &Captioning & 35.56 & 37.54 & 38.59 & 39.32 & 39.65 & 9.90 & 5.72 & 3.12 & 1.34 & 0.69 & 43.16 & 48.95 & 52.83 & 56.26 & 58.24 \\
& &Fusion & \textbf{51.70} & \textbf{54.65} & \textbf{56.23} & \textbf{57.25} & \textbf{57.67} & \textbf{14.38} & \textbf{8.39} & \textbf{4.63} & \textbf{2.01} & \textbf{1.04} & \textbf{62.29} & \textbf{70.84} & \textbf{76.62} & \textbf{81.32} & \textbf{83.67} \\
& &Reranking & 42.37 & 45.02 & 46.74 & 48.40 & 49.42 & 11.81 & 6.91 & 3.89 & 1.77 & 0.97 & 51.84 & 59.59 & 66.00 & 73.86 & 79.67 \\
\hline\hline
\multirow{2}{*}{\textbf{Text-only}}
& \multirow{2}{*}{\textbf{Text-only}}
& Random     & 0.01 & 0.01 & 0.02 & 0.05 & 0.08 & 0.00 & 0.00 & 0.00 & 0.00 & 0.00 & 0.01 & 0.03 & 0.06 & 0.19 & 0.41 \\
& & \textbf{Others}*  & 50.71 & 53.39 & 54.74 & 55.60 & 55.97 & 14.11 & 8.10 & 4.39 & 1.87 & 0.96 & 61.41 & 69.21 & 74.19 & 78.23 & 80.37 \\

\midrule
\multirow{5}{*}{\textbf{Text-only}}
& \multirow{5}{*}{\textbf{All}}
& Random     & 0.01 & 0.01 & 0.02 & 0.04 & 0.07 & 0.00 & 0.00 & 0.00 & 0.00 & 0.00 & 0.02 & 0.02 & 0.05 & 0.15 & 0.30 \\
& & Text-only  & 48.98 & 51.71 & 53.15 & 54.06 & 54.45 & 13.68 & 7.90 & 4.31 & 1.84 & 0.95 & 59.78 & 67.72 & 73.04 & 77.31 & 79.54 \\
& & Captioning & 50.45 & 53.15 & 54.51 & 55.40 & 55.76 & 14.03 & 8.07 & 4.38 & 1.86 & 0.96 & 61.12 & 68.99 & 74.01 & 78.16 & 80.29 \\
& & Fusion     & 50.70 & 53.39 & 54.73 & 55.60 & 55.96 & 14.11 & 8.10 & 4.39 & 1.87 & 0.96 & 61.41 & 69.23 & 74.18 & 78.25 & 80.40 \\
& & Reranking  & 48.93 & 51.67 & 53.11 & 54.02 & 54.40 & 13.67 & 7.90 & 4.31 & 1.84 & 0.95 & 59.72 & 67.70 & 73.01 & 77.32 & 79.53 \\
\midrule\midrule

\multirow{5}{*}{\textbf{Multimodal}} 
& \multirow{5}{*}{\textbf{Multimodal}}
& Random & 0.06 & 0.09 & 0.12 & 0.21 & 0.31 & 0.02 & 0.02 & 0.02 & 0.02 & 0.02 & 0.09 & 0.17 & 0.30 & 0.74 & 1.32 \\
& & Text-only & 8.70 & 9.82 & 10.93 & 12.29 & 13.20 & 2.44 & 1.60 & 1.04 & 0.57 & 0.35 & 11.12 & 14.50 & 18.77 & 25.43 & 30.83 \\
& & Captioning & 5.22 & 5.90 & 6.47 & 7.25 & 7.81 & 1.44 & 0.95 & 0.60 & 0.33 & 0.21 & 6.68 & 8.72 & 10.94 & 14.74 & 18.11 \\
& & Fusion & 55.07 & 58.55 & 60.70 & 62.07 & 62.57 & 15.34 & 9.19 & 5.27 & 2.38 & 1.25 & 65.52 & 75.57 & 83.29 & 89.47 & 92.15 \\
& & Reranking & 27.79 & 30.25 & 32.66 & 36.04 & 38.56 & 7.66 & 4.75 & 2.97 & 1.64 & 1.04 & 34.33 & 41.63 & 50.67 & 66.69 & 81.04 \\\hline

\multirow{5}{*}{\textbf{Multimodal}}
& \multirow{5}{*}{\textbf{All}}
& Random     & 0.01 & 0.01 & 0.01 & 0.03 & 0.06 & 0.00 & 0.00 & 0.00 & 0.00 & 0.00 & 0.01 & 0.03 & 0.04 & 0.13 & 0.28 \\
& & Text-only  & 4.19 & 5.12 & 6.10 & 7.22 & 7.99 & 1.34 & 0.99 & 0.71 & 0.41 & 0.26 & 6.15 & 8.95 & 12.73 & 18.27 & 22.86 \\
& & Captioning & 1.16 & 1.49 & 1.80 & 2.17 & 2.44 & 0.35 & 0.29 & 0.21 & 0.12 & 0.08 & 1.69 & 2.72 & 3.90 & 5.73 & 7.34 \\
& & Fusion     & 53.88 & 57.38 & 59.54 & 60.92 & 61.48 & 14.96 & 9.00 & 5.19 & 2.34 & 1.24 & 64.09 & 74.23 & 82.00 & 88.14 & 91.21 \\
& & Reranking  & 27.44 & 29.81 & 32.21 & 35.54 & 38.09 & 7.53 & 4.64 & 2.91 & 1.61 & 1.03 & 33.77 & 40.84 & 49.83 & 65.65 & 80.20 \\ 

\bottomrule
\end{tabular}
}
\end{table*}

\begin{table*}[t]
\centering
\caption{Retrieval performance (\%) on ~{\DS-M} dataset under different retrieval settings. }
\vspace{-0.1in}
\label{tab:medmkg_retrieval}
\resizebox{1\textwidth}{!}{%
\begin{tabular}{c|c|l|ccccc|ccccc|ccccc}
\toprule
\multirow{2}{*}{\textbf{\shortstack{Query\\ Type}}}
& \multirow{2}{*}{\textbf{\shortstack{Triplet\\Type}}}
& \multirow{2}{*}{\textbf{Method}}
& \multicolumn{5}{c|}{\textbf{NDCG @$K$ $\uparrow$}}
& \multicolumn{5}{c|}{\textbf{Precision @$K$ $\uparrow$}}
& \multicolumn{5}{c}{\textbf{Recall @$K$ $\uparrow$}} \\
\cline{4-18}
& & & 5 & 10 & 20 & 50 & 100
& 5 & 10 & 20 & 50 & 100
& 5 & 10 & 20 & 50 & 100 \\
\midrule
\multirow{5}{*}{\textbf{All}}
& \multirow{5}{*}{\textbf{All}}
& Random & 0.00 & 0.01 & 0.02 & 0.04 & 0.09 & 0.00 & 0.01 & 0.01 & 0.01 & 0.01 & 0.01 & 0.03 & 0.06 & 0.13 & 0.34 \\
& & Text-only & 24.72 & 28.19 & 30.43 & 32.25 & 33.06 & 14.48 & 9.88 & 6.12 & 2.96 & 1.63 & 27.93 & 36.31 & 42.86 & 49.24 & 52.58 \\
& & Captioning & 24.60 & 28.02 & 30.24 & 32.06 & 32.88 & 14.38 & 9.80 & 6.07 & 2.94 & 1.62 & 27.74 & 36.01 & 42.50 & 48.93 & 52.33 \\
& & Fusion & 26.53 & 30.17 & 32.68 & 34.85 & 35.93 & 15.62 & 10.63 & 6.62 & 3.25 & 1.82 & 29.67 & 38.55 & 45.84 & 53.46 & 57.88 \\
& & Reranking & \textbf{27.85} & \textbf{32.48} & \textbf{36.60} & \textbf{41.36} & \textbf{42.25} & \textbf{16.82} & \textbf{12.01} & \textbf{8.12} & \textbf{4.50} & \textbf{2.41} & \textbf{31.36} & \textbf{42.42} & \textbf{54.01} & \textbf{70.03} & \textbf{73.66} \\ \midrule\midrule

\multirow{2}{*}{\textbf{Text-only}}
& \multirow{2}{*}{\textbf{Text-only}}
& Random     & 0.03 & 0.03 & 0.07 & 0.13 & 0.20 & 0.02 & 0.02 & 0.02 & 0.03 & 0.02 & 0.02 & 0.04 & 0.14 & 0.38 & 0.70 \\
& & \textbf{Others}*  & 41.93 & 47.81 & 51.61 & 54.69 & 56.07 & 24.56 & 16.75 & 10.38 & 5.02 & 2.76 & 47.37 & 61.59 & 72.69 & 83.51 & 89.17 \\
\midrule

\multirow{5}{*}{\textbf{Text-only}}
& \multirow{5}{*}{\textbf{All}}
& Random     & 0.01 & 0.01 & 0.01 & 0.03 & 0.06 & 0.01 & 0.00 & 0.00 & 0.01 & 0.01 & 0.00 & 0.00 & 0.01 & 0.08 & 0.24 \\
& & Text-only  & 41.93 & 47.80 & 51.61 & 54.69 & 56.06 & 24.56 & 16.75 & 10.38 & 5.02 & 2.76 & 47.37 & 61.59 & 72.69 & 83.51 & 89.17 \\
& & Captioning & 41.71 & 47.51 & 51.26 & 54.33 & 55.70 & 24.39 & 16.61 & 10.28 & 4.98 & 2.74 & 47.05 & 61.07 & 72.01 & 82.84 & 88.53 \\
& & Fusion     & 41.93 & 47.80 & 51.61 & 54.69 & 56.06 & 24.56 & 16.75 & 10.38 & 5.02 & 2.76 & 47.37 & 61.59 & 72.69 & 83.51 & 89.17 \\
& & Reranking  & 41.81 & 47.64 & 51.42 & 54.50 & 55.87 & 24.48 & 16.68 & 10.32 & 5.00 & 2.75 & 47.18 & 61.33 & 72.35 & 83.16 & 88.84 \\
\midrule\midrule
\multirow{5}{*}{\textbf{Multimodal}}
& \multirow{5}{*}{\textbf{Multimodal}}
 & Random & 0.01 & 0.01 & 0.02 & 0.06 & 0.14 & 0.01 & 0.00 & 0.01 & 0.02 & 0.02 & 0.01 & 0.01 & 0.02 & 0.16 & 0.49 \\
& & Text-only & 0.00 & 0.00 & 0.00 & 0.00 & 0.00 & 0.00 & 0.00 & 0.00 & 0.00 & 0.00 & 0.00 & 0.00 & 0.00 & 0.00 & 0.00 \\
& & Captioning & 0.17 & 0.19 & 0.25 & 0.31 & 0.39 & 0.10 & 0.07 & 0.06 & 0.03 & 0.03 & 0.22 & 0.28 & 0.43 & 0.65 & 1.02 \\
& & Fusion & 4.41 & 4.83 & 5.48 & 6.36 & 7.01 & 2.79 & 1.83 & 1.23 & 0.71 & 0.45 & 4.25 & 5.47 & 7.27 & 10.28 & 12.92 \\
& & Reranking & 7.81 & 10.71 & 15.32 & 22.48 & 22.67 & 5.81 & 5.30 & 4.94 & 3.79 & 1.93 & 8.62 & 15.25 & 27.67 & 51.16 & 51.85 \\ \hline
\multirow{5}{*}{\textbf{Multimodal}}
& \multirow{5}{*}{\textbf{All}}
& Random     & 0.03 & 0.05 & 0.05 & 0.08 & 0.10 & 0.02 & 0.02 & 0.01 & 0.01 & 0.01 & 0.03 & 0.06 & 0.07 & 0.17 & 0.27 \\
& & Text-only  & 0.00 & 0.00 & 0.00 & 0.00 & 0.00 & 0.00 & 0.00 & 0.00 & 0.00 & 0.00 & 0.00 & 0.00 & 0.00 & 0.00 & 0.00 \\
& & Captioning & 0.01 & 0.01 & 0.04 & 0.06 & 0.09 & 0.01 & 0.00 & 0.01 & 0.01 & 0.01 & 0.01 & 0.01 & 0.10 & 0.21 & 0.32 \\
& & Fusion     & 4.41 & 4.83 & 5.48 & 6.36 & 7.01 & 2.79 & 1.83 & 1.23 & 0.71 & 0.45 & 4.25 & 5.47 & 7.27 & 10.28 & 12.92 \\
& & Reranking  & 7.81 & 10.71 & 15.32 & 22.48 & 22.67 & 5.81 & 5.30 & 4.94 & 3.79 & 1.93 & 8.62 & 15.25 & 27.67 & 51.16 & 51.85 \\ 

\bottomrule
\end{tabular}
}
\end{table*}

\section{Benchmark Evaluation}

\subsection{Evaluation Settings}
\label{sec:settings}
We evaluate the proposed {\DS} on both retrieval and generation tasks. As our primary focus is on the \textbf{training-free} setting, all evaluations are conducted exclusively on the test splits reported in Table~\ref{tab:mmkg_rt_stats}, without using any training or validation data.
As described in Section~\ref{sec:data_curation} (Step~3), the benchmark includes two types of queries—text-only and multimodal—and the underlying multimodal knowledge graphs contain both text-only and multimodal triplets\footnote{Note that MarKG and MedMKG do not include triplets of the form (\textit{image, relation, image}).}. To comprehensively assess retrieval and generation performance under different modality combinations, we evaluate the following five settings, defined by the query modality and the scope of candidate triplets:
(\textit{S1}) \textbf{all queries with all triplets},
(\textit{S2}) \textbf{text-only queries with text-only triplets},
(\textit{S3}) \textbf{text-only queries with all triplets},
(\textit{S4}) \textbf{multimodal queries with multimodal triplets}, and
(\textit{S5}) \textbf{multimodal queries with all triplets}.

Setting S1 provides an overall evaluation of different retrieval techniques under the most general and realistic scenario. Comparing S2 and S3 allows us to assess whether incorporating multimodal triplets benefits retrieval and generation for text-only queries. Similarly, comparing S4 and S5 evaluates the contribution of text-only triplets when handling multimodal queries. Together, these settings enable systematic analysis of how multimodal information influences retrieval and generation performance across different query types and knowledge graph configurations.

\subsection{Retrieval Evaluation}
\label{sec:retr_eval}
\noindent \textbf{Evaluation Configuration.}
The objective of retrieval evaluation is to assess the effectiveness of different retrieval techniques introduced in Section~\ref{sec:pre_baseline}, including \textit{text-only retrievers}~\cite{hu2025graggraphretrievalaugmentedgeneration, li2023graph, he2024g}, \textit{fusion-based multimodal retrievers}~\cite{chen2022murag, multimodalSurvey}, \textit{captioning-based retrievers}~\cite{Gao2022, SelfAdaptive}, and \textit{reranking-based retrievers}~\cite{Yan2024, LLMreranker}. In addition, we include a basic \textit{random retriever} as a simple lower-bound baseline for the retrieval task.
For fair comparison, all retrievers are implemented using a shared CLIP encoder~\cite{radford2021learning}\footnote{\url{https://huggingface.co/sentence-transformers/clip-ViT-B-32}} to obtain unified representations. Additionally, the \textit{captioning-based retrievers} are implemented with a BLIP model~\cite{li2022blip}\footnote{\url{https://huggingface.co/Salesforce/blip-image-captioning-base}}. Both queries and candidate triplets are embedded into the same representation space, and candidates are ranked based on \textbf{cosine similarity}. We report standard retrieval metrics, including NDCG$@K$, Precision$@K$, and Recall$@K$ in the main experiments. Implementation details are provided in Appendix~\ref{app:retr_imple}.

\medskip
\noindent \textbf{Retrieval Results and Key Findings.}
We evaluate retrieval performance on both {\DS}-G and {\DS}-M under different query and triplet compositions. Detailed results are reported in Table~\ref{tab:markg_retrieval} and Table~\ref{tab:medmkg_retrieval}, with efficiency analysis provided in Appendix~\ref{app:efficiency}. Note that \textbf{Others*} means the baselines mentioned in the evaluation configuration.
Rather than enumerating numerical comparisons, we summarize the results through a set of key findings that highlight the challenges and opportunities of multimodal KG-RAG retrieval.

\ul{\textbf{Finding 1:} \textit{A modality gap persists in retrieval.}}
For text-only queries retrieving from text-only triplets, all retrievers exhibit nearly identical performance, as they effectively reduce to the same text-based retrieval pipeline. When the candidate space is expanded to include both text-only and multimodal triplets, performance changes only marginally, and top-ranked results remain dominated by text-only triplets. This observation suggests that multimodal triplets rarely serve as effective matches for purely textual queries and primarily increase the size of the candidate pool without contributing useful evidence.
More broadly, this behavior reveals a persistent \emph{modality gap} in the retrieval space: cross-modal matching between textual queries and multimodal knowledge does not emerge naturally from hybrid indexing alone. This finding indicates that simply incorporating multimodal knowledge into a KG is insufficient for enabling cross-modal retrieval, and that explicit mechanisms are required to bridge modalities when such interactions are desired.

\ul{\textbf{Finding 2:} \textit{Multimodal embedding is critical for visually grounded retrieval.}}
Once queries require visual grounding, unimodal retrieval strategies become unreliable. Text-only retrievers degrade sharply because they lack the ability to incorporate image evidence. Captioning-based retrievers also underperform, as the captioning process introduces an information bottleneck: omissions or inaccuracies in generated captions directly propagate into retrieval errors.
In contrast, embedding-based multimodal retrievers remain effective in this regime. Fusion-based methods benefit from jointly encoding visual and textual signals into a shared representation space, enabling more robust cross-modal alignment. Reranking-based approaches further improve performance by refining candidate sets using stronger multimodal matching. These results indicate that the central capability for effective MKG-RAG retrieval is not improved text ranking, but reliable multimodal representation learning that preserves and aligns visual evidence with KG semantics.

\ul{\textbf{Finding 3:} \textit{Retrieval difficulty and method preference are domain-dependent.}}
Visually grounded retrieval is consistently more challenging on {\DS}-M than on {\DS}-G. A plausible explanation is that medical images tend to be visually homogeneous and require sensitivity to subtle patterns, while the associated terminology is dense and fine-grained. As a result, generic similarity signals are weaker and more easily confounded.
This domain discrepancy also explains differences in method preference. On ~{\DS}-G, fusion-based retrieval is often sufficient, as coarse semantic alignment yields strong performance. On ~{\DS}-M, reranking-based approaches provide greater benefits, as refinement helps disambiguate subtle visual cues and specialized language. These observations highlight that retrieval strategies effective in general domains may not transfer directly to specialized domains without adaptation.

\ul{\textbf{Implications for MKG-RAG retriever design.}}
Together, these findings suggest several directions for future retriever development. First, bridging the modality gap requires explicit cross-modal alignment objectives or supervision, rather than relying on mixed-modality indexing to induce cross-modal matching implicitly. Second, multimodal retrieval should prioritize representation learning that preserves fine-grained visual information and aligns it directly with KG semantics, as lossy caption-based conversions consistently limit performance. Third, domain-aware refinement becomes increasingly important in settings with subtle visual distinctions and specialized terminology, motivating stronger reranking or verifier-style matching modules and the use of domain-adapted visual and textual encoders. Overall, effective MKG-RAG retrieval is likely to require the integration of explicit cross-modal alignment, robust multimodal embeddings, and domain-sensitive refinement.

\begin{table*}[t]
\centering
\caption{Generation results (\%) on ~{\DS}-G and ~{\DS}-M with $K=5$.}
\vspace{-0.1in}
\label{tab:rag_em_f1_contains_gm}
\resizebox{0.9\textwidth}{!}{%
\begin{tabular}{c|c|l|cccc|cccc}
\toprule
\multirow{2}{*}{\textbf{\shortstack{Query\\Type}}}
& \multirow{2}{*}{\textbf{\shortstack{Triplet\\Type}}}
& \multirow{2}{*}{\textbf{Method}}
& \multicolumn{4}{c|}{\textbf{\DS-G}}
& \multicolumn{4}{c}{\textbf{\DS-M}} \\
\cline{4-11}
& &
& \textbf{EM} $\uparrow$ & \textbf{F1} $\uparrow$ & \textbf{Contains}$@1$ $\uparrow$ & \textbf{BLEU} $\uparrow$
& \textbf{EM} $\uparrow$ & \textbf{F1} $\uparrow$ & \textbf{Contains}$@1$ $\uparrow$ & \textbf{BLEU} $\uparrow$ \\
\midrule

\multirow{6}{*}{\textbf{All}}
& \multirow{6}{*}{\textbf{All}}
& RAG-free    & 5.04  & 15.15  & 16.50  & 43.03  & 1.34  & 7.84  & 4.51  & 44.72 \\
& & Random      & 8.99  & 17.82 & 18.76 & 47.27 & 1.67  & 7.91  & 6.68  & 50.61 \\
& & Text-only   & 13.62 & 25.50 & 26.72 & 50.63 & 13.69 & 25.80 & 22.70 & 54.35 \\
& & Captioning  & 13.20 & 25.12 & 26.66 & 50.55 & 14.69 & 25.76 & 22.20 & 55.19 \\
& & Fusion      & \textbf{16.76} & \textbf{29.96} & \textbf{31.77} & \textbf{53.03} & 13.86 & 25.53 & 23.87 & 55.86 \\
& & Reranking   & 14.47 & 26.99 & 28.63 & 51.42 & \textbf{16.86} & \textbf{28.78} & \textbf{26.88} & \textbf{56.78} \\
\midrule\midrule

\multirow{3}{*}{\textbf{Text-only}}
& \multirow{3}{*}{\textbf{Text-only}}
& RAG-free         & 4.64  & 15.00  & 16.05  & 41.20  & 0.85  & 9.35  & 4.25  & 45.15 \\
& & Random           & 8.65  & 17.69 & 18.18 & 46.10 & 3.40  & 12.39 & 11.33 & 54.79 \\
& & \textbf{Others}* & 15.55 & 28.13 & 29.63 & 51.29 & 21.46 & 38.72 & 38.53 & 61.77 \\
\midrule

\multirow{5}{*}{\textbf{Text-only}}
& \multirow{5}{*}{\textbf{All}}
& Random      & 9.36  & 17.76 & 18.45 & 46.51 & 1.47  & 10.38 & 8.50  & 53.37 \\
& & Text-only   & 15.91 & 28.30 & 29.17 & 51.59 & 21.99 & 38.22 & 36.95 & 61.63 \\
& & Captioning  & 15.89 & 28.26 & 29.41 & 51.47 & 23.46 & 38.40 & 36.36 & 62.47 \\
& & Fusion      & 16.38 & 28.69 & 29.78 & 51.80 & 21.41 & 37.80 & 36.95 & 61.69 \\
& & Reranking   & 15.94 & 28.06 & 28.98 & 51.32 & 23.46 & 38.18 & 36.66 & 61.96 \\
\midrule\midrule

\multirow{6}{*}{\textbf{Multimodal}}
& \multirow{6}{*}{\textbf{Multimodal}}
& RAG-free    & 5.52  & 14.79  & 16.19  & 46.57  & 2.03  & 5.78  & 2.85  & 45.50 \\
& & Random      & 7.85  & 18.36 & 20.90 & 49.86 & 2.44  & 5.53  & 4.07  & 49.05 \\
& & Text-only   & 8.28  & 19.61 & 21.66 & 49.97 & 1.63  & 4.28  & 2.44  & 45.32 \\
& & Captioning  & 8.34  & 21.19 & 24.85 & 50.48 & 2.44  & 4.94  & 3.25  & 43.58 \\
& & Fusion      & 18.46 & 34.35 & 38.06 & 56.73 & 3.25  & 10.26 & 6.50  & 49.14 \\
& & Reranking   & 11.42 & 24.49 & 27.40 & 51.88 & 7.32  & 15.19 & 12.60 & 47.80 \\
\midrule

\multirow{5}{*}{\textbf{Multimodal}}
& \multirow{5}{*}{\textbf{All}}
& Random      & 8.14  & 17.96 & 19.47 & 49.03 & 1.94  & 4.65  & 4.26  & 46.96 \\
& & Text-only   & 8.36  & 19.09 & 21.09 & 48.41 & 2.71  & 9.38  & 3.88  & 44.73 \\
& & Captioning  & 7.01  & 17.89 & 20.33 & 48.44 & 3.10  & 9.05  & 3.49  & 45.57 \\
& & Fusion      & 17.64 & 32.87 & 36.35 & 55.86 & 3.88  & 9.32  & 6.59  & 48.15 \\
& & Reranking   & 11.11 & 24.55 & 27.83 & 51.64 & 8.14  & 16.37 & 13.95 & 49.93 \\
\bottomrule
\end{tabular}
}
\vspace{-0.1in}
\end{table*}

\subsection{Generation Evaluation}

\noindent \textbf{Evaluation Configuration.}
For generation evaluation, we include a \textit{generation-only} baseline (i.e., RAG-free) that queries the LLM without retrieval augmentation, and evaluate RAG variants built on the same retrievers, modality settings, and implementations described in Section~\ref{sec:retr_eval}. For each query, we provide the top-$K$ retrieved triplets as additional context to the generator, using $K{=}5$ to balance evaluation fidelity with computational cost and token budget. We use GPT-5~\cite{singh2025openai} as the generation model. Generation is evaluated as an open-ended (visual) question answering task using Exact Match (EM), token-level F1, Contains$@1$, and BLEU-1.

\smallskip
\noindent \textbf{Generation Results and Key Findings.}
Table~\ref{tab:rag_em_f1_contains_gm} reports generation performance on ~{\DS}-G and ~{\DS}-M under different query and knowledge modality combinations. Rather than enumerating individual scores, we summarize the results through several key findings that reveal how retrieval quality and modality interact to shape end-to-end MKG-RAG performance.

\ul{\textbf{Finding 4:} \textit{Retrieval provides genuine but uneven utility for generation.}}
Compared to the generation-only baseline, retrieval-augmented generation yields consistent improvements across most metrics and settings, confirming that retrieved multimodal knowledge can be effectively consumed by the generator. Notably, even Random retrieval occasionally improves performance, indicating that when relevant triplets appear in the context, the generator can exploit them. This serves as a sanity check for the alignment between constructed knowledge, queries, and answers, and validates the overall quality of the benchmark.

\ul{\textbf{Finding 5:} \textit{Generation gains are substantially weaker in multimodal regimes.}}
Retrieval augmentation delivers the largest gains when both queries and supporting knowledge are textual. In contrast, for multimodal queries, improvements are often marginal unless a strong multimodal retriever is used, and even then absolute gains remain limited—particularly on ~{\DS}-M. These results indicate that end-to-end generation is highly sensitive to whether retrieval can reliably supply visually grounded evidence. Simply incorporating multimodal knowledge into the pipeline does not automatically translate into proportional generation improvements, revealing a key bottleneck in current MKG-RAG systems.

\ul{\textbf{Finding 6:} \textit{Generation performance closely tracks retrieval quality.}}
Generation trends closely mirror retrieval behavior. Text-only and captioning-based retrieval pipelines provide limited gains for multimodal queries because they either ignore visual information or compress it through lossy captions, constraining evidence quality before generation begins. In contrast, embedding-based multimodal retrievers yield the most reliable improvements in multimodal settings. Fusion-based methods are generally more effective on ~{\DS}-G, while reranking-based approaches offer additional benefits on {\DS}-M, where subtle visual distinctions and specialized terminology demand stronger refinement. This alignment highlights retrieval quality as a primary driver of end-to-end MKG-RAG generation performance.

\ul{\textbf{Implications: toward graph-aware multimodal generation.}}
The comparatively modest gains in multimodal settings suggest that current multimodal retrievers, largely designed for unstructured corpus retrieval, do not fully exploit the structure and semantics of MKG. Improving MKG-RAG generation, therefore, likely requires retrieval methods that are explicitly graph-aware, such as relation-sensitive matching, neighborhood- or path-based evidence aggregation, and graph-constrained candidate selection to reduce distractors. Beyond retrieval, there is also headroom in how evidence is presented to the generator, including improved triplet selection, ordering, and representation formats that better convey structured graph information. Overall, these findings suggest that progress in MKG-RAG will depend less on scaling generic multimodal retrieval and more on developing graph-aware, domain-sensitive retrieval and evidence integration strategies.




\begin{table*}[t]
\centering
\caption{Generation results (\%) for validating the effectiveness of LLM-based utility filtering strategy.}
\vspace{-0.1in}
\label{tab:ablation}
\begin{tabular}{c|l|cccc}
\toprule
\textbf{Split} & \textbf{Method} &
\textbf{EM} $\uparrow$ & \textbf{F1} $\uparrow$ & \textbf{Contains@1} $\uparrow$ & \textbf{BLEU} $\uparrow$ \\
\midrule
\multirow{2}{*}{\textbf{Filtered}} 
& RAG-free & 3.33 & 11.09 & 8.00  & 41.35 \\
& Fusion   & 8.00 & 14.04 & 12.00 & 46.21 \\
\midrule
\multirow{2}{*}{\textbf{Remaining}} 
& RAG-free & 5.52  & 14.79 & 16.19 & 46.57 \\
& Fusion   & 18.46 & 34.35 & 38.06 & 56.73 \\

\bottomrule
\end{tabular}%
\end{table*}

\begin{table}[t]
\centering

\caption{Cross-LLM consistency analysis.}
\vspace{-0.1in}
\label{tab:llm_dependency}
\begin{tabular}{lcc}
\toprule
\textbf{Model} & \textbf{Filter Alignment (\%)} & \textbf{BERTScore} \\
\midrule
Qwen3.5 & 99.7 & 0.8818 \\
Gemini 2.5 Flash & 97.0 & 0.8818 \\
\bottomrule
\vspace{-0.15in}
\end{tabular}
\end{table}

\subsection{Ablation on LLM-based Utility Filtering}
\label{app:ablation}

A key design choice in {\DS} is to filter out low-utility triplets that tend to produce ambiguous or weakly grounded queries. To validate this choice, we conduct an ablation study comparing questions derived from \textbf{filtered} triplets and those derived from the \textbf{remaining} triplets retained in the benchmark. We randomly sample 150 multimodal questions from each group and evaluate MKG-RAG performance under Setting (\textit{S4}) using our strongest Fusion-based retriever, with both RAG-free and retrieval-augmented generation. Results are reported in Table~\ref{tab:ablation}.
Across all metrics, questions derived from filtered triplets consistently yield lower performance than those derived from remaining triplets, even without retrieval augmentation, indicating that such questions are intrinsically difficult and poorly specified. Although retrieval improves performance in both cases, gains on filtered-triplet questions are substantially smaller, suggesting that retrieval cannot compensate for unclear or weakly grounded information needs. These results demonstrate that low-utility triplets confound reliable MKG-RAG evaluation, and therefore validate the necessity of LLM-based utility filtering to ensure clear, grounded queries and faithful assessment of retrieval and generation performance.

\subsection{Discussion on LLM Dependency}
\label{sec:llm_dependency}

To evaluate whether the pipeline depends heavily on a specific LLM, we conduct a cross-LLM consistency analysis. Besides the GPT-based construction pipeline, we apply Qwen3.5~\cite{qwen35blog} and Gemini 2.5 Flash~\cite{comanici2025gemini25} to a 300-sample subset, including 150 text-only samples and 150 multimodal samples. We compare their outputs with GPT-based results using filtering alignment and question-level semantic similarity measured by BERTScore. As shown in Table~\ref{tab:llm_dependency}, both alternative LLMs achieve high filtering alignment with GPT-based outputs, with Qwen3.5 reaching 99.7\% and Gemini 2.5 Flash reaching 97.0\%. The generated questions also show strong semantic consistency, with a BERTScore of 0.8818 for both models. These results indicate that the filtering and synthesis process is robust to the choice of LLM backbone.

We further conduct a human evaluation to assess the reliability of the filtering and question synthesis results. Human annotators reach 88\% agreement with the filtering decisions, and the generated questions receive an average quality and diversity score of 3.97 out of 5. Together, the cross-LLM and human evaluation results show that the benchmark construction process maintains stable behavior across different LLMs and introduces limited model-specific bias in practice.

%% file: tex/conclusion.tex
\section{Conclusion}
This work demonstrates that progress in MKG-RAG is increasingly constrained by retrieval, a component that has received limited attention largely due to the lack of suitable benchmarks. To address this gap, we introduce \DS, a cross-domain benchmark explicitly designed to evaluate retrieval in MKG-RAG, comprising two multimodal knowledge graphs and carefully aligned question-answering datasets. {\DS} is constructed through an LLM-based curation pipeline that filters low-utility knowledge and systematically covers diverse modality configurations, enabling controlled and realistic retrieval evaluation. Extensive experiments across representative retriever families reveal that effectively leveraging multimodal signals during retrieval remains both challenging and critical for end-to-end performance. By isolating and diagnosing these challenges, {\DS} provides a principled foundation for developing, evaluating, and advancing more effective retrievers for MKG-RAG. We hope this benchmark will catalyze future research on graph-aware, domain-adaptive multimodal retrieval methods and more effective integration of structured knowledge into multimodal generation.

%% file: tex/appendix.tex

\begin{figure*}[t]
\centering

\begin{tcolorbox}[
  title={Utility Filtering Prompt},
  colback=white,
  colframe=blue!50!black,
  boxrule=0.8pt,
  arc=2pt
]
\small
You curate and write natural-language queries for a KG retrieval dataset.\par
Modality: \{multimodal or text-only\}.\par
At query time, the head is given as: \{IMAGE or TEXT\}.\par
\vspace{0.6\baselineskip}

Task: This group corresponds to \{FIXED (head, relation) with MULTIPLE possible tails \textbf{OR} FIXED (head, tail) with MULTIPLE possible relations\}.\par
You must decide if this group is meaningful for generating questions that ask for the missing part.\par
\vspace{0.6\baselineskip}


\textbf{Rules:}\par
1) Output MUST be JSON: \verb|{"keep": bool, "reason": str}|.\par
2) Return ONLY JSON (no extra keys).\par
3) Decide keep based primarily on relation\_text and tail examples.\par
4) Because head\_text may be uninformative for images, DO NOT reject just because head\_text is generic.\par
5) Set keep=false only when the tails are mostly placeholders or administrative/meta concepts, e.g.,\par
tails like `thing/object/entity/space/place/environment/location/category', or when the relation is too vague.\par
\vspace{0.6\baselineskip}

\textbf{Input:}\par
GROUP TYPE: \{mask\_tail or mask\_relation\}\par
head\_text: \{head\_text\}\par
relation\_text: \{relation\_text\}\par
tail\_text (if mask\_relation): \{tail\_text\}\par
candidates\_count: \{cand\_total\}\par
examples (random 3): \{cand\_examples\}\par
candidates\_sample: \{cand\_sample\}\par
\end{tcolorbox}

\vspace{6pt}

\begin{tcolorbox}[
  title={Question Generation Prompt},
  colback=white,
  colframe=green!50!black,
  boxrule=0.8pt,
  arc=2pt
]
\small
You curate and write natural-language queries for a KG retrieval dataset.\par
Modality: \{multimodal or text-only\}.\par
At query time, the head is given as: \{IMAGE or TEXT\}.\par
\vspace{0.6\baselineskip}

Task: This group corresponds to \{FIXED (head, relation) with MULTIPLE possible tails \textbf{OR} FIXED (head, tail) with MULTIPLE possible relations\}.\par
Generate exactly K WH-questions that ask for the missing part.\par
\vspace{0.6\baselineskip}

\textbf{Rules:}\par
1) Output MUST be JSON: \verb|{"questions": [..]}|.\par
2) \{prefix\_rule\}\par
3) \{head\_leak\_rule\}\par
4) Do NOT copy relation\_text (or candidate relation strings) verbatim; paraphrase.\par
5) If mask\_tail: Questions MUST NOT mention any specific tail candidates.\par
6) $\le$ 25 words per question. Avoid yes/no.\par
7) Refer to the head as `\{head\_ref\}' (or equivalent if multimodal).\par
8) If mask\_relation: You MAY mention tail\_text.\par
\vspace{0.6\baselineskip}

\textbf{Input:}\par
GROUP TYPE: \{mask\_tail or mask\_relation\}\par
head\_text: \{head\_text\}\par
relation\_text: \{relation\_text\}\par
tail\_text (if mask\_relation): \{tail\_text\}\par
candidates\_count: \{cand\_total\}\par
examples (random 3): \{cand\_examples\}\par
candidates\_sample: \{cand\_sample\}\par
K=\{k\}\par
\end{tcolorbox}

\vspace{-8pt}
\caption{Two-agent prompting for query construction: filtering low-quality triplets and generating WH-questions for the remaining triplets.}
\label{fig:agent_prompt}
\vspace{-0.1in}
\end{figure*}

\section{Source Knowledge Graphs}
\label{app:source}
We use the following two knowledge graphs:
\begin{itemize}[leftmargin=*]
    \item \textbf{MarKG.}
MarKG~\cite{zhang2023multimodalanalogicalreasoningknowledge} is a multimodal knowledge graph constructed to support multimodal analogical reasoning, where the entities involved in an analogy (e.g., head/tail/query entities) may come from different modalities. It is released together with the Multimodal Analogical Reasoning dataset (MARS) and is built by expanding and annotating seed entities and relations from E-KAR~\cite{chen2022kar} and BATS~\cite{drozd2016word}. In particular, MarKG links external entities to Wikidata and associates entities with images collected from LAION-5B~\cite{schuhmann2022laion5bopenlargescaledataset}. Overall, MarKG contains 11,292 entities and 192 relation types, forming 34,420 triplets, with 76,424 linked images, providing multimodal grounding for general-domain analogy.

\item \textbf{MedMKG.}
For the multimodal knowledge graph, we use MedMKG~\cite{wang2025medmkgbenchmarkingmedicalknowledge}. MedMKG is a multimodal knowledge graph that fuses rich imaging data from MIMIC-CXR~\cite{Johnson2019-fk} with structured clinical knowledge from the Unified Medical Language System (UMLS)~\cite{Bodenreider2004-aq}, using both rule-based tools and large language models for precise concept extraction and relationship modeling. To enhance graph quality and compactness, Neighbor-aware Filtering (NaF), a filtering algorithm specifically designed for multimodal knowledge graphs, is applied to MedMKG. MedMKG comprises 53,587 edges, 4,868 images, and 262 relations over 3,148 concepts.
\end{itemize}

\begin{table*}[t]
\centering
\caption{Retrieval efficiency measured by Queries per Second (QPS) under an online serving setting with cached embeddings and warm-started retrievers. We conduct the experiments with all settings as detailed in Section~\ref{sec:settings}.}
\vspace{-0.1in}
\label{tab:qps}
\begin{tabular}{c|l|ccccc}
\toprule
\textbf{Dataset}&
\textbf{Retriever} & \textbf{S1} & \textbf{S2} & \textbf{S3} & \textbf{S4} & \textbf{S5} \\
\midrule
\multirow{4}{*}{\textbf{\DS-G}}
& \textbf{Text-only}   & 108.96 & 108.17 & 109.42 &  95.41 & 107.28 \\
& \textbf{Captioning}  & 102.72 & 108.95 & 107.05 &  97.99 & 108.53 \\
& \textbf{Fusion}      &  20.82 & 116.42 & 117.11 &  20.73 &  20.97 \\
& \textbf{Reranking}   &  21.91 & 114.44 & 121.12 &  22.46 &  23.60 \\
\midrule
\multirow{4}{*}{\textbf{\DS-M}}
& \textbf{Text-only}   &  69.68 &  69.99 &  69.53 &  69.58 &  69.54 \\
& \textbf{Captioning}  &  79.52 &  77.00 &  79.32 &  72.92 &  79.41 \\
& \textbf{Fusion}      &   7.89 &  78.81 &  77.24 &   7.72 &   7.84 \\
& \textbf{Reranking}   &   7.94 &  77.23 &  76.83 &   7.72 &   7.65 \\
\bottomrule
\end{tabular}%
\end{table*}

\section{Prompts for Data Construction}
\label{app:prompts}

During {\DS} construction, we prompt GPT-5 twice: first to validate knowledge utility, and then to translate triplets into natural-language queries. The corresponding prompts are shown in Figure~\ref{fig:agent_prompt}.

\section{Image Augmentation for Visual Questions}
\label{app:image}

To prevent exact image duplication from serving as a shortcut in multimodal retrieval, we apply image augmentation to the images used in visual questions. For each question image, we randomly apply one augmentation from the following set: \textit{swapping to another image in the knowledge graph that refers to the same entity, random cropping, rotation, or color jitter}. These augmentations ensure that the image in the visual question differs from the image associated with the target triplet, eliminating trivial matches and better simulating realistic retrieval settings where an exact duplicate of the query image is unlikely to appear in the knowledge graph.

\section{Example of {\DS} Construction}
\label{app:cons_example}


We use two examples to illustrate how the proposed pipeline generates benchmark data with high-utility triplets.
\begin{itemize}[leftmargin=*]
    \item \textbf{Text-only Query.} We take (\textit{Antigua and Barbuda, member\_of, [MASK]}) as example. The pipeline will generate a query first, which is \ul{``\textit{Of which global or intergovernmental bodies is Antigua and Barbuda a member?}''} For the \textbf{retrieval} task, the ground truth data include several high-utility triples, such as \{(\textit{Antigua and Barbuda, member\_of, Commonwealth of Nations}), (\textit{Antigua and Barbuda, member\_of, United Nations}), (\textit{Antigua and Barbuda, member\_of, World Trade Organization}). For the \textbf{generation} task, the answers are \{\textit{Commonwealth of Nations, United Nations, World Trade Organization}\}.
    \item \textbf{Multimodal Query.} If there is a high-utility multimodal triplet (\textit{People's Republic of China} \includegraphics[height=1.em]{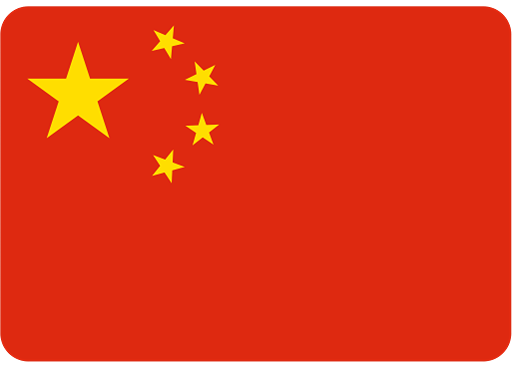}\textit{, railway\_traffic\_side, left}), the entity ``\textit{left}'' is then masked, leading to the generation of the multimodal query: \includegraphics[height=1.em]{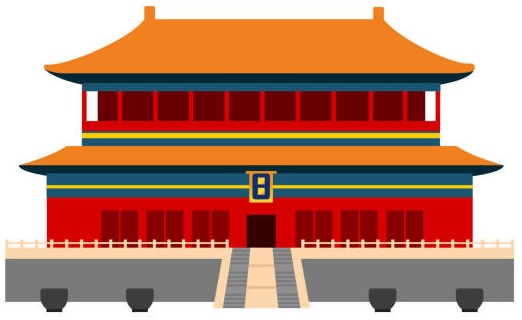} \ul{``\textit{On which side do trains operate in the subject’s country?}''} Note that the image entities can be different. For the \textbf{retrieval} task, the ground triplet is the original high-utility triplet, and for the \textbf{generation} task, the answer is the mask ``\textit{left}''.

\end{itemize}







\section{Details about Retriever Implementation}
\label{app:retr_imple}

The proposed {\DS} is a \textbf{general} benchmark, and any retrievers can be used. However, in the current version, we implement the following retrievers in the experiments.

For text-only queries and triplets, all retrievers that share the same CLIP implementation produce identical embeddings by encoding text solely with CLIP’s text encoder. As a result, they achieve the same performance on text-only query–text-only triplet retrieval due to this homogeneous encoding.
For multimodal retrieval, the text-only retriever ignores visual signals and encodes only the textual entity associated with the image. The captioning-based retriever first uses BLIP to convert the image into a textual description, and then encodes the resulting text with CLIP’s text encoder.

The remaining two retrievers adopt more advanced multimodal strategies. We implement the fusion retriever using a late-fusion design, where the visual embedding from CLIP’s image encoder is averaged with the text embedding of the textual part of each multimodal triplet. The reranking retriever performs a two-stage procedure: it first retrieves candidates using image-embedding similarity, and then reranks the shortlisted triplets using text-embedding similarity.

\section{Retrieval Efficiency Analysis}
\label{app:efficiency}

An additional study analyzes the retrieval efficiency of the baseline retrievers. We report Queries per Second (QPS) under an online serving setting where all embeddings are cached and the retrievers are warm-started. The results are shown in Table~\ref{tab:qps}. Overall, the findings highlight a clear \textbf{efficiency--effectiveness trade-off}: in the multimodal setting, fusion-based and reranking-based retrievers exhibit roughly an order-of-magnitude lower throughput (about 0.1$\times$ QPS) than the remaining baselines. By contrast, captioning-based retrieval benefits substantially from caching and achieves competitive throughput, approaching that of the text-only retriever. Taken together, this study quantifies the efficiency cost of stronger multimodal modeling and provides practical guidance for deploying MKG-RAG systems under real-time constraints.